\documentclass[10pt,twocolumn,letterpaper]{article}

\usepackage[pagenumbers]{cvpr} 

\usepackage{graphicx}
\usepackage{amsmath}
\usepackage{amssymb}
\usepackage{booktabs}
\usepackage{bm}
\usepackage{mathtools}
\usepackage{algpseudocode}
\usepackage{algorithm}
\usepackage{siunitx}
\usepackage{multirow}
\usepackage{enumitem}
\usepackage{caption}
\usepackage{diagbox}
\usepackage[dvipsnames]{xcolor}
\usepackage[accsupp]{axessibility}

\definecolor{myGreen}{RGB}{0,114,0}

\usepackage[pagebackref,breaklinks,colorlinks,citecolor=myGreen]{hyperref}

\usepackage[capitalize]{cleveref}
\crefname{section}{Sec.}{Secs.}
\Crefname{section}{Section}{Sections}
\Crefname{table}{Table}{Tables}
\crefname{table}{Tab.}{Tabs.}

\def\cvprPaperID{8666} 
\def\confName{ICCV}
\def\confYear{2023}

\graphicspath{ {images/} }

\algnewcommand\algorithmiclocalize{\textbf{Localize neuron and update model:}}
\algnewcommand\Localize{\item[\algorithmiclocalize]}
\algnewcommand\algorithmiccross{\textbf{Cross-section plane determination:}}
\algnewcommand\Cross{\item[\algorithmiccross]}
\algnewcommand\algorithmicbranchc{\textbf{Bifurcation candidates detection:}}
\algnewcommand\Branchc{\item[\algorithmicbranchc]}

\newcommand{\diag}[1]{\text{diag}}
\newcommand{\conj}[1]{\text{conj}}

\newcommand{\independent}{\perp \!\!\! \perp}

\makeatletter

\usepackage{algorithmicx,algorithm}
\definecolor{commentcolor}{RGB}{110,154,155}   
\newcommand{\PyComment}[1]{\ttfamily\textcolor{commentcolor}{\# \small #1}}  
\newcommand{\PyCommentNewLine}[1]{\ttfamily\textcolor{commentcolor}{~~~~\# \small #1}}  
\newcommand{\PyCode}[1]{\ttfamily\textcolor{black}{ \small #1}} 
\definecolor{functioncolor}{RGB}{200,2,127}   
\newcommand{\PyFunc}[1]{\ttfamily\textcolor{functioncolor}{\small #1}}  

\begin{document}

\title{Causal-DFQ: Causality Guided Data-free Network Quantization}

\author{
  \textbf{Yuzhang Shang$^{1,2}$, Bingxin Xu$^1$, Gaowen Liu$^{2}$, Ramana Rao Kompella$^{2}$,}
  \textbf{Yan Yan$^{1}$\thanks{Corresponding author}}\\
  $^{1}$Illinois Institute of Technology, $^{2}$Cisco Research \vspace{-3pt}\\
\tt\small{\{yshang4, bxu21\}@hawk.iit.edu, \{yuzshang, gaoliu, rkompell\}@cisco.com, yyan34@iit.edu}
}
\maketitle

\begin{abstract}
Model quantization, which aims to compress deep neural networks and accelerate inference speed, has greatly facilitated the development of cumbersome models on mobile and edge devices. There is a common assumption in quantization methods from prior works that training data is available. In practice, however, this assumption cannot always be fulfilled due to reasons of privacy and security, rendering these methods inapplicable in real-life situations. 
Thus, data-free network quantization has recently received significant attention in neural network compression. Causal reasoning provides an intuitive way to model causal relationships to eliminate data-driven correlations, making causality an essential component of analyzing data-free problems. However, causal formulations of data-free quantization are inadequate in the literature. To bridge this gap, we construct a causal graph to model the data generation and discrepancy reduction between the pre-trained and quantized models. Inspired by the causal understanding, we propose the Causality-guided Data-free Network Quantization method, Causal-DFQ, to eliminate the reliance on data via approaching an equilibrium of causality-driven intervened distributions. Specifically, we design a content-style-decoupled generator, synthesizing images conditioned on the relevant and irrelevant factors; then we propose a discrepancy reduction loss to align the intervened distributions of the pre-trained and quantized models. It is worth noting that our work is the first attempt towards introducing causality to data-free quantization problem. Extensive experiments demonstrate the efficacy of Causal-DFQ. The code is available at \href{https://github.com/42Shawn/Causal-DFQ}{Causal-DFQ}.
\end{abstract}
\section{Introduction}
\label{sec:intro}
There have been significant advances in deep learning models in the fields of computer vision~\cite{he2016deep,girshick2015fast}  and natural language processing~\cite{sutskever2014sequence,pennington2014glove}. To accommodate the increasing demand for equipping cumbersome models on resource-constrained edge devices, researchers have proposed several network quantization methods~\cite{zhou2016dorefa,hubara2016binarized}, in which high-precision parameters are converted into low-precision ones. To mitigate the performance degradation induced by model quantization, fine-tuning approaches are extensively studied to optimize quantized models on the full training datasets~\cite{jacob2018quantization,wu2016quantized,shang2022lipschitz,shang2022network}. However, original training data is sometimes inaccessible in real-world situations due to the privacy and security concerns. A patient's electronic health record, for instance, is typically inaccessible because the information contained is private. Hence, the fine-tuning methods requiring training data are no longer applicable in such real-life scenarios.

To address this issue, researchers have proposed data-free quantization to quantize models without requiring access to real data~\cite{banner2018aciq,cai2020zeroq,xu2020generative,zhang2021diversifying,liu2021data,shang2023post}. For example, ZeroQ \cite{cai2020zeroq} is proposed to generate `optimal' fake data, which learns an input data distribution to best match the batch normalization statistics of the FP32 model. Nevertheless, most data-free quantization methods attempt to reconstruct the original data from the pre-trained model utilizing prior statistical distribution information of the underlying data, such as BNS~\cite{cai2020zeroq,xu2020generative,yin2020dreaming}, Dirichlet distribution~\cite{nayak2019zero} and category information~\cite{chen2019data}. However, those methods ignore a powerful tool in the human cognition, \ie, causal reasoning, which commonly aids humans in learning without relying upon data collection. Human cognitive systems are immune to the data deficiency because humans are more sensitive to causal relations than data-driven statistical associations~\cite{gopnik2004theory,zhang2021adversarial}. Using causal language, causal reasoning can extract causal relationship from the pre-trained models and ignore irrelevant factors by interventions~\cite{peters2017elements}. 

There are two significant challenges that need to be overcome before causality can be introduced to eliminate the reliance on data during the quantized model training. First, constructing an informative causal graph is the fundamental premise for causal reasoning~\cite{pearl2009causality,peters2017elements}, but how causal graphs should be constructed in a data-free situation is still inadequate in the literature.
Second, using causal language to formalize data generation and network alignment is the key to connecting causality with data-free quantization, but it also remains unsolved. These two challenges are the fundamental obstacles that prevent us from employing causality in data-free quantization.

To address these challenges, we construct a causal graph to model the data-free quantization process, including data-generation and discrepancy reduction mechanisms, where the irrelevant factors in the pre-trained models are taken into consideration. Based on the causal graph, we propose a novel Causality-Guided Data Free Network Quantization method, \textbf{\emph{Causal-DFQ}}, to remove the reliance on data during quantized model training. Specifically, we design a content-style-decoupled generator, synthesizing images conditioned on the relevant and irrelevant factors (content and style variables). Then we propose a discrepancy reduction loss to align the intervened distributions of the outputs from pre-trained and quantized models. 

Overall, the contributions of this paper are four-fold:
    \textbf{(i)} We provide a causal perspective on data-free quantization, which is the first attempt towards using causality to facilitate data-free network compression;
    \textbf{(ii)} To leverage causality to facilitate data-free quantization, we construct a causal graph to model data generation process and discrepancy reduction process in data-free quantization mechanism;
    \textbf{(iii)} We propose a novel quantization method called Causality Guided Data-free Network Quantization, \emph{Causal-DFQ}, in which we generate fake images conditioned on style and content variables, and align style-intervened distributions of pre-trained and quantized models.
    \textbf{(iv)} Extensive experiments demonstrate that the proposed method can significantly improve the performance of data-free low-bit models. Importantly, it is the first method where data-free fine-tuned models outperform the models fine-tuned with data on the ImageNet.
\section{Related Work}
\label{related}

\noindent\textbf{Data-free Network Compression.}
Although model compression has become a hot topic recently, compressing model without training data still is a challenge. As pioneers, \cite{srinivas2015data} initially devise a channel pruning method without original training data. And then a large number of data-free (DF) or zero-shot compression methods were proposed, \eg DF quantization~\cite{banner2018aciq,cai2020zeroq,xu2020generative,zhang2021diversifying,liu2021data}, DF factorization~\cite{nagel2019data} and DF knowledge distillation~\cite{lopes2017data,chen2019data,fang2019data}. Especially for DF quantization, recent work~\cite{banner2018aciq,cai2020zeroq,xu2020generative,zhang2021diversifying,liu2021data} go further to data-free quantization, which requires neither training nor validation data for quantization. Most of the data-free KD methods attempt to reconstruct the original data from the pre-trained model utilizing prior information about the underlying data statistical distribution, such as BNS~\cite{cai2020zeroq,xu2020generative,yin2020dreaming}, Dirichlet distribution~\cite{nayak2019zero} and category information~\cite{chen2019data}. However, all existing methods overlook causal reasoning, a powerful tool for humans to cognize even in situations where data are inaccessible.

\noindent\textbf{Causal Reasoning.}
\label{related_causal}
One core purpose of causal reasoning is to pursue the causal effect of interventions, contributing to achieving the desired objective. Recent work shows the benefits of introducing causality into machine learning from various aspects~\cite{scholkopf2021toward}. After the deep connections of causal systems and the concept of exogeneity having been successfully implemented in social science, such as in Economics and Genetics~\cite{pearl2009causality}, Sch{\"o}lkopf \textit{et al.}~\cite{scholkopf2012causal} originally develop a technique, named independence mechanisms via introducing causal mechanisms to independently separate the exogenous and endogenous variables \textit{w.r.t.} specific tasks in the field of machine learning~\cite{scholkopf2021toward}. 

However, thanks to the unique nature of data-free quantization, our data generation process is steerable, unlike previous works. 
Thus, we can design a content-style-decoupled generator where both content and style variables are accessible in the causal graph. Then we can easily implement do-calculus~\cite{pearl2012calculus} for causal reasoning.

\section{Method}
In this section, we elaborate on the methodology of Causality-guided Data-Free Quantization, named \emph{Causal-DFQ}. Firstly, we review the idea of network quantization and the general framework of data-free compression. Secondly, we construct the causal graph model for the data-free network compression, which is adopted as a theoretical tool to bridge causality with data-free quantization. Thirdly, based on the causal graph, we observe that there is a unique property of data-free compression, where the data variable is completely accessible; thus we design a generator to synthesize images conditioned on style and content images for training quantized models. Next, we focus on the optimization formulation that converts the causal task into an optimizable problem. Finally, we discuss the potential insights for the \emph{Causal-DFQ}. 
Note that we only elaborate on the key derivations in this section due to the space limitation. Detailed discussions, technical theorems, and implementation details in Codes can be found in the supplemental materials.

\subsection{Preliminary}
Here, we revisit the basic ideas of network quantization and data-free network compression. 

\noindent\textbf{Network Quantization.}
Network quantization is a popular technique for compressing neural networks. The quantization function is the key to train a neural network with low-precision weights and activations. 
The most common quantization function is called uniform quantization function, which is pioneerly proposed in~\cite{zhou2016dorefa}. The uniform quantization function $q(\cdot)$ for $k$-bit quantization is defined as follows: 
\begin{equation}
    q(\textit{v}) = \text{round}(L\cdot(\textit{v}-Z)),
\end{equation}
where \textit{v} denotes a scalar value (full-precision, float32), $L$ is the scaling factor, and $Z$ is the zero point in float32. According to whether the parameter $Z$ is zero, uniform quantization can be categorized into symmetric quantization and asymmetric quantization. In our work, we use symmetric quantization, \ie, $Z = 0$, and then $S$ is written as follows:
\begin{equation}
    S=\frac{2^{k-1}-1}{\max(\vert x_f\vert)},
\end{equation}
where $x_f$ is the full-precision numbers.

\noindent\textbf{Data-Free Compression.}
The key of most network compression methods~\cite{gou2021knowledge,hubara2016binarized} including quantization is to reduce the discrepancy $\mathcal{D}$ between the pre-trained full-precision model $f(\cdot;\theta_P)$
and the quantized model with low-precision weights 
$f(\cdot;\theta_Q)$ 
through optimizing 
$\theta_Q$. The idea of discrepancy reduction can be formalized as follows:
\begin{equation}
    f(\cdot;\theta_Q)^{\star}=\min_{\theta_Q}\mathcal{D}(f(\cdot;\theta_P),f(\cdot;\theta_Q)).
\end{equation}
This discrepancy reduction module can be considered as a knowledge distillation mechanism for aligning model $f(\cdot;\theta_P)$ and $f(\cdot;\theta_Q)$. Consequently, the integral framework of the data-free quantization can be considered as the incorporation of a generator and knowledge distillation~\cite{hinton2015distilling,shang2021lipschitz}, and the core idea is to reconstruct some samples from full-precision models to fine-tune quantized models~\cite{liu2021zero,liu2021data}. 
Therefore, to achieve the goal of data-free compression via causality, we modify the existing framework from the perspectives of generator and discrepancy reduction.

\subsection{Causal Graph of Data-Free Compression}
Humans can perform causal reasoning, an essential ability that makes humans learn differently from machine learning algorithms. The superiority of causal reasoning endows humans with the ability to identify causal relationships. This allows them to ignore irrelevant factors that are not causally related to the targeted task and removes the reliance on collecting data for learning~\cite{peters2017elements,zhang2020causal,scholkopf2021toward,zhang2021adversarial}. Contrary to this, neural networks are normally trained based on data-driven correlation. In other words, neural networks do not have the ability to distinguish causal relationships. 
In the absence of this ability, irrelevant factors in data are overfitted, further resulting in the overreliance of networks on data. For instance, in a car recognition case, road background is a data-driven irrelevant factor yet cannot reflect the causality $\wrt$ the targeted task, \ie, human can recognize a car with causal reasoning even if it is not on the road. Moreover, data even are unavailable in our data-free case. Therefore, we desire to incorporate causal reasoning to remove the reliance on data, \ie, the pre-trained model guides the training of the quantized model via causality in a data-free manner.
Before performing causal reasoning to guide the training in a data-free manner, we need to construct a causal graph since causal graphs are the key to formulating causal reasoning~\cite{peters2017elements,zhang2021adversarial}. 
In the context of data-free quantization, we desire a causal graph by which the distributions of the outputs of pre-trained and quantized models can be included. Besides, the graph is expected to reflect the impact of irrelevant factors on these two output distributions, and then we are able to align the distributions. Specifically, we investigate the difference in irrelevant factors between these two distributions and enforce the quantized models to focus on the relevant factors. Consequently, this encourages the quantized model to learn via causal reasoning. 

There are two general approaches to building a causal graph of the targeted learning mechanism. One approach is to use causal structure learning to infer causal graphs ~\cite{pearl2009causality,peters2017elements,scholkopf2021toward}, but it is challenging to apply this approach to high-dimensional data. Using external knowledge to construct causal graphs is another approach~\cite{tang2020long,scholkopf2021toward,zhang2021adversarial}. As automatically learning a precise causal graph is out of scope for this work, external human knowledge of the data generation process is employed to construct the causal graph. Here, we aim to construct a causal graph to model the data-free quantization process, including data-generation and discrepancy reduction mechanisms, where the irrelevant factors in the pre-trained models are also considered. 

\begin{figure}[t!]
  \begin{center}
  \vspace{-0.1in}
    \includegraphics[width=0.45\textwidth]{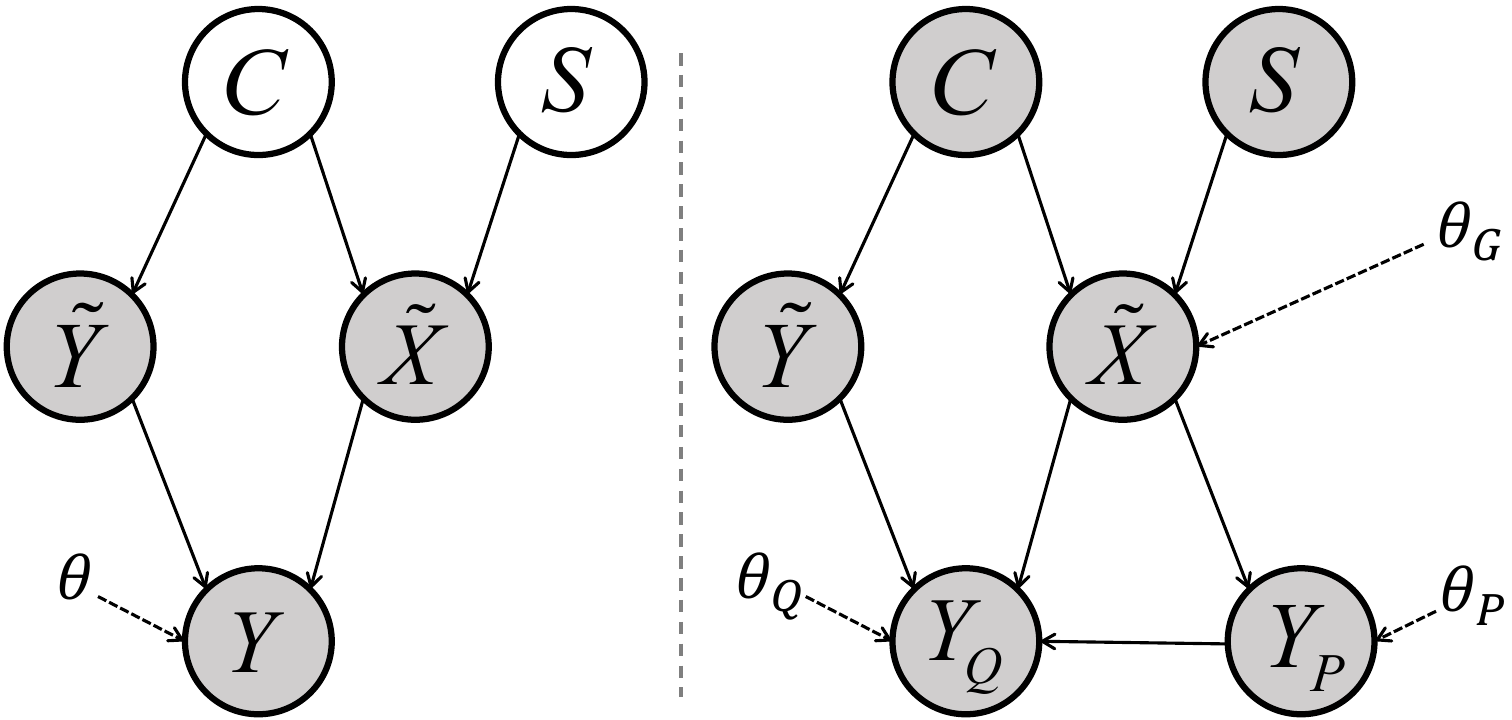}
  \end{center}
  \caption{\textbf{Left}: Causal graph of the ideal data generation and model learning process. \textbf{Right}: Causal graph of the data-free quantization process. Each node represents a random variable, and shallow ones indicate observable variables, where $C$, $S$, $\tilde{X}$, $\tilde{Y}$, $Y_P$, $Y_Q$, $\theta_G$, $\theta_P$, $\theta_Q$ are content variable, style variable, generated data, generated label, distilled label, output label, parameters of the generator, parameters of the pre-trained model, parameters of the quantized, respectively.}
  \label{fig:causal_graph}
  \vspace{-0.2in}
\end{figure}

Specifically, we construct a causal graph $\mathcal{G}$ to formalize the general idea of data-free quantization process including an image generation mechanism and a discrepancy reduction mechanism to allow the pre-trained model $f(\cdot;\theta_P)$ to guide the training of the quantized model $f(\cdot;\theta_Q)$. 
In the previous studies~\cite{peters2017elements,scholkopf2021toward,zhang2021adversarial}, even though there is a number of different causes of natural data, researchers ideally and effectively divide all the causes into two categories for simplicity. We follow the existing work, and group content-related causes into one category, called content variable $C$. The rest causes, \ie, irrelevant factors, are grouped into another category, called style variable $S$, which is content-independent, \ie, $S\independent C$. This implies that $C \rightarrow \tilde{X} \leftarrow  S$ and $C \rightarrow \tilde{Y}$. Then the generated data are fed into the pre-trained model and quantized model. Under the supervision of the output of pre-trained model $Y_P$ and the generated label $\tilde{Y}$, we obtain the output of quantized model $Y_Q$. The causal graph is shown in Fig.~\ref{fig:causal_graph}. Based on the causal graph, we first use a structural causal model~\cite{pearl2000models} to represent the data generating mechanism:
\begin{equation}
    \tilde{X}:= \mathcal{M}(S,C,\theta_G),
    \label{eq:data_generation}
\end{equation}
where $\theta_G$ is the parameters of generator. 

After formulating the process of obtaining the generated data, we expect to define valid interventions and the corresponding intervention distributions~\cite{pearl2009causality,scholkopf2021toward}. Defining valid interventions is equivalent to determining which variables or mechanisms in the causal graph can be intervened. The common practice is to utilize the independence mechanism~\cite{scholkopf2012causal} to \textit{construct probabilistic relations in causal reasoning} and \textit{discover the irrelevant factors as the intervened variable}. This practice has been proven to be an effective way to realize causality reasoning by previous work~\cite{scholkopf2012causal,scholkopf2019causality,huang2020causal,mitrovic2020representation}, in which the conditional (intervened) distribution does not change under the interventions on irrelevant variable (\ie, $S$).  Theoretically, for an image generation mechanism (ideally even collected natural images are included), $P(\tilde{Y}\mid C)$ is invariant to $S$. Therefore, we claim $C$ as a representation of invariant content of data \emph{w.r.t.} the $\tilde{Y}$ under interventions $I$ on style domain $\mathcal{S}$ as shown Fig.~\ref{fig:causal_graph}(left), and the relationship can be mathematically denoted as:
\begin{equation}
    P^{do(S=i_l)}(\tilde{Y} \mid C) = P^{do(S=i_k)}(\tilde{Y} \mid C) 
    ~~~\forall i_l, i_k \in \mathcal{I},
    \label{eq:general_traget}
\end{equation}
where $i_l$ and $i_k$ form a pair of interventions in the domain of interventions $\mathcal{I}$, and $P^{do(S=i_l)}$ stands for the distribution under intervention $i_l$ on $\mathcal{S}$~\cite{pearl2009causality}. In the data-free compression literature, we also desire to access the intervened distributions of the outputs of the pre-trained model and quantized model and then derive a computationally reachable equilibrium between them. 
\textit{Because our data generation process is steerable, unlike the fixed datasets collected from natural distributions, we design a content-style-decoupled generator where both content and style variables are accessible in the causal graph.}

Therefore, based on the analysis of Eq.\ref{eq:data_generation}, \ref{eq:general_traget} and the above-mentioned nature of the data-free mechanism, the desirable equilibrium in data-free quantization can be formulated as follows: $\forall l, k \in \left\{1,2,\cdots,M\right\}$,
\begin{equation}
\begin{aligned}
    P^{do(S=i_l)}(Y_P\mid f(\tilde{X};\theta_Q)) = P^{do(S=i_k)}(Y_P\mid f(\tilde{X};\theta_Q)),
\end{aligned}
\end{equation}
which can be reformed as follows:
\noindent\textbf{Targeted Causal Equilibrium:~~}
\begin{equation}
\begin{aligned}
     &P^{do(S=i_l)}(f(\tilde{X};\theta_P)\mid f(\tilde{X};\theta_Q))\\
    =&P^{do(S=i_k)}(f(\tilde{X};\theta_P)\mid f(\tilde{X};\theta_Q)),
\end{aligned}
\label{eq:realgoal}
\end{equation}
where $M$ is the number of interventions in style domain $\mathcal{S}$, $f(\cdot;\theta_P)$ and $f(\cdot;\theta_Q)$ are the pre-trained and quantized model with parameters $\theta_P$ and $\theta_Q$, respectively. 
Straight-forwardly, we desire the distribution, $P(f(\tilde{X};\theta_P)\mid f(\tilde{X};\theta_Q))$ to be invariant over style variable change.


\subsection{Content-Style-Decoupled Generator}

Here, we present the design of the content-style-decoupled generator, rendering accessibility to the content and style variables. Structural Equation Modeling (SEM)~\cite{wright1949genetical,pearl2000models} is a primary causal model, which is originally proposed to apply explicit causal interpretations to regression equations based on direct and indirect effects of observed variables in the fields of Genetics and Economics~\cite{pearl2009causality}. SEM has two main components: the structural model showing potential causal dependencies between endogenous and exogenous variables and the measurement model showing the relations between latent variables and their indicators. SEM aims to obtain an informative representation of some observable output. 

\begin{figure*}[ht]
    \centering
    \includegraphics[width=0.9\textwidth]{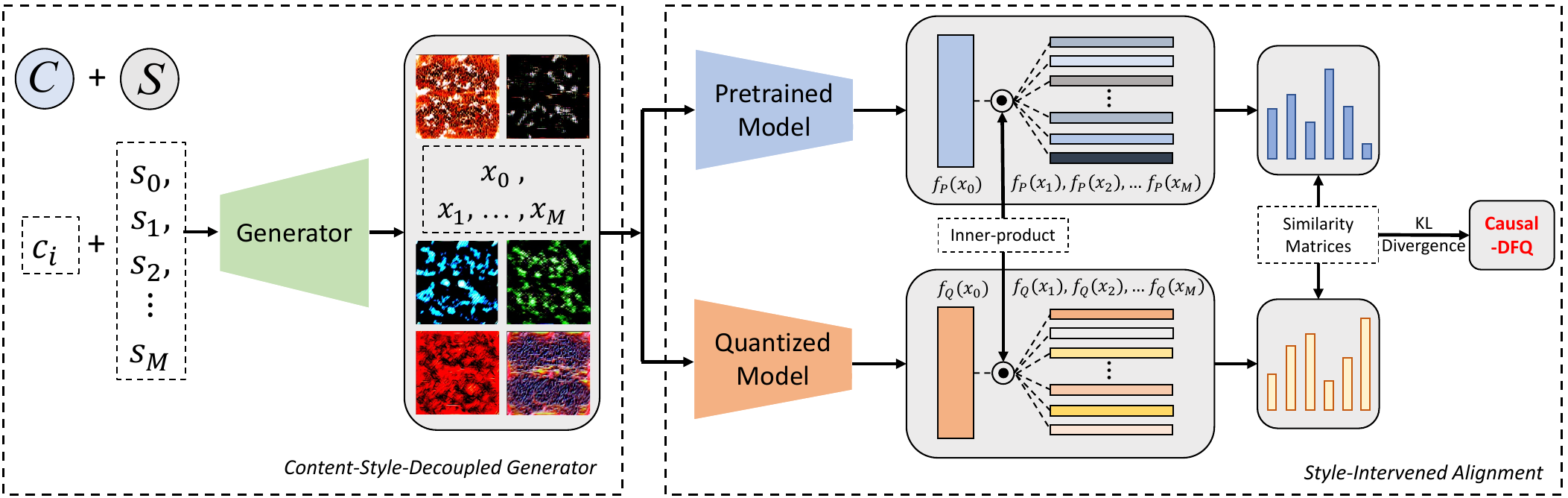}
    \captionsetup{font=small}
    \caption{\textbf{Overview of the pipeline.}
    To eliminate the reliance on data and utilize the causality in the discrepancy reduction stage, we disentangle the invariant content $C$ and semantics-irrelevant style $S$ in the view of causality. We propose a \textbf{Content-Style-Decoupled Generator} to synthesize fake images conditioned on the independent content and style variables. Follow by the generator, we design \textbf{\emph{Causal-DFQ}} loss to achieve knowledge exclusively based on content by intervening with the style variable. In particular, we use KL-divergence to minimize the distance between conditional distributions (similarity matrices, calculated in a contrastive way) of pre-trained and quantized models.
}
    \label{pipeline:2}
    \vspace{-0.2in}
\end{figure*}

Inspired by the concepts of SEM, we naturally introduce the above two variables into the data-free scenarios and expect to enforce the outputs of quantized models exclusively correlated with the content variable in the view of causal reasoning. To realize the goal, we generalize three fundamental assumptions of SEM into the literature on data-free compression~\cite{pearl2012calculus,scholkopf2012causal,mitrovic2020representation}. The generalized assumptions can be interpreted as follows: 
\textbf{(i)} The data are generated from content variables $C$ representing factors inside the model and style variables $S$ (irrelevant factors outside the model) for targeted tasks. \textbf{(ii)} Only variable $C$ is relevant for the model output, \textit{i.e.}, content dominates the model performance. \textbf{(iii)} Content and style are causally independent, \textit{i.e.}, style changes are content-preserving. 

\begin{algorithm}[!t]
    \PyFunc{def} \PyCode{ generator(S, C):} \PyComment{generator}\\    
    \PyCode{~~~~input = torch.mul(Embedding(C), S)} \\    
    \PyCommentNewLine{style \& content fusion} \\
    \PyCode{~~~~x = conv\_blocks(input)} \\    
    \PyCommentNewLine{generate images via conv layers} \\
    \PyFunc{~~~~return} \PyCode{x} \\
    \PyCode{content = torch.randint(0, class\_number, (batch\_size),))} \PyComment{define content}\\
    \PyCode{style = torch.randn(batch\_size, latent\_dim)} \\
    \PyCommentNewLine{define style}\\
    \PyCode{generated\_x = generator(style, content)} \\
    \PyCommentNewLine{generate images based on C and S}
\caption{Pseudo code of Content-Style-Decoupled Generator in a PyTorch-like style.}
\label{algo:generator}
\end{algorithm}

Based on the above assumptions, to achieve the goal of directly performing interventions on the style domain and building the equilibrium of the intervened distributions as presented in Eq.~\ref{eq:realgoal}, we design an image generator that can synthesize fake data conditioned on independent style and content variables. We call this generator a content-style-decoupled generator. Specifically, we assign every to-be-generated sample a content label and a style noise; then we feed this pair of content and style into the generator network to generate the sample. In this way, for each sample of the following discrepancy reduction process, we can access its style variable and perform interventions by keeping its content labels consistent and adjusting its style noise. 

Here, we give a straightforward explanation of how our generator produces fake images conditioned on content and style variables (Algorithm~\ref{algo:generator}). First, the integer function, \texttt{content = randint()} (function of randomly generating non-nagetive integer) generates the pseudo label based on the number of classes of the real dataset, which can be interpreted as a content variable for each generated image. Note that using the number of classes does not imply information leakage and is still within our data-free scenarios, as we can acquire the number of classes via accessing the pre-trained model's classification head rather than accessing the labels. And the Gaussian noise generation function, \texttt{style = randn()} can assign a Gaussian noise to style variable. By pairing the content and style and then feeding them to a generator, we can synthesize fake data conditioned on the content and style. In this way, we can directly manipulate the style variable. More details can be found in the \emph{codes in the Supplemental Materials}.

\subsection{Style-Intervened Discrepancy Reduction}

After accessing the style variable in the data-free quantization mechanism, the only remaining problem is to achieve the equilibrium of the intervened distributions as derived in Eq.~\ref{eq:realgoal}. 
We maintain the invariance under interventions via a regularization term to address this. The optimization problem is formalized as follows:
\begin{equation}
\begin{aligned}
    \min \displaystyle \mathop{\mathbb{E}}_{X\in \mathcal{D}} \mathop{\mathbb{E}}_{\{i_{lk},i_{qt}\}}\Bigl[\mathcal{L}_{i_{lk}}&(f(X;\theta_P), f(X;\theta_Q)) \\
   + \mathcal{L}_{i_{qt}}&(f(X;\theta_P), f(X;\theta_Q))\Bigr].\\
    \textit{s.t.} ~~~ KL\Bigl[ P^{do(S=i_{lk})}&(f(X;\theta_P) \mid f(X;\theta_Q)), \\ P^{do(S=i_{qt})}&(f(X;\theta_P) \mid f(X;\theta_Q))\Bigr]\leq \tau
\end{aligned}
\label{eq:objective}
\end{equation}
where $i_{lk} \triangleq i_l \times i_k \sim\mathcal{I}\times\mathcal{I}$ stands for a pair of interventions, $\mathcal{L}$ is the vanilla alignment loss, and $KL(\cdot,\cdot)$ is the KL-divergence. $\tau$ is a small threshold to adjust the similarity between two distributions. Any distance measure on distributions can be used in place of the KL divergence such as cross-entropy, since we only expect the intervened distributions $P^{do(S=i_{lk})}(f(X;\theta_P) \mid f(X;\theta_Q))$ and $P^{do(S=i_{qt})}(f(X;\theta_P) \mid f(X;\theta_Q))$ to be similar. 
In practice, we define the output representations of pre-trained and quantized models (\textit{i.e.}, $f(X;\theta_P)$ and $f(X;\theta_Q)$) at the penultimate layer.

How to approach the conditional distribution under interventions $P^{do(S=i_{lk})}(f(\tilde{X};\theta_P) \mid f(\tilde{X};\theta_Q))$ becomes the key problem. To estimate the distribution, we introduce the noise-contrastive estimation (NCE)~\cite{gutmann2010noise,hjelm2018learning}. Specifically, we take pairs of points $(x_i, x_j)$ to compute similarity scores and use pairs of intervention $i_{lk}$ to perform a style intervention. Given a batch of samples $\left\{x_i\right\}, i \in \left\{1,2,\cdots,N\right\}$, the conditional probability of the pair can be estimated as follows:
\begin{equation}
\begin{aligned}
    P^{do(S=i_{lk})}(f(X;\theta_P) \mid f(X;\theta_Q)) \\
    \propto h(f(x^{S=i_l}_{j};\theta_P), f(x^{S=i_{k}}_{i};\theta_Q)),
\end{aligned}
\label{eq:approximatation}
\end{equation}
in which $h$ is the function to measure the similarity between the representations of the pre-trained model $f(x^{S=i_l}_{j};\theta_P)$ and the one from quantized model $f(x^{S=i_k}_{i};\theta_Q)$. Using this function to estimate the conditional distribution is originally proposed in NCE~\cite{gutmann2010noise}, also called the critic in contrastive learning~\cite{hjelm2018learning}. It is defined as below:
\begin{equation}
\begin{aligned}
    h(\mathbf{x}, \mathbf{y}) = \exp(\frac{<g(\mathbf{x}), g(\mathbf{y})>}{\beta}),
\end{aligned}
\label{eq:critic}
\end{equation}
where $\beta$ is the temperature to adjust degree of concentration, and $g$ is a fully-connected network~\cite{gutmann2010noise}. 

Combining all the equations, we obtain the optimizable objective function as follows: $\mathcal{L}_{\emph{Causal-DFQ}} = $
\begin{equation}
\begin{aligned}
    \displaystyle \mathop{\mathbb{E}}_{X\in \mathcal{D}} \mathop{\mathbb{E}}_{\left\{i_{lk},i_{qt}\right\}} & \Bigr[\mathcal{L}_{i_{lk}}(f(X;\theta_P), f(X;\theta_Q)) \\
    &+ \mathcal{L}_{i_{qt}}(f(X;\theta_P), f(X;\theta_Q))\Bigl]\\
    +\sum_{i_{lk}}\sum_{i_{qt}} & KL\Bigl[ P^{do(S=i_{lk})}(f(X;\theta_P) \mid f(X;\theta_Q)), \\
    &P^{do(S=i_{qt})}(f(X;\theta_P) \mid f(X;\theta_Q))\Bigr].
\end{aligned}
\label{eq:optimization}
\end{equation}
Concretely, the probability of a pair of samples in the conditional distribution $P^{do(S=i_{lk})}(f(X;\theta_P) \mid f(X;\theta_Q))$ can be approximated by the critic function as follows~\cite{gutmann2010noise,hjelm2018learning,mitrovic2020representation}:
\begin{equation}
\begin{aligned}
  &P^{do(S=i_{lk})}(f(x_j;\theta_P) \mid f(x_i;\theta_Q)) \\
  =& \frac{h(f(x^{S=i_l}_{j};\theta_P), f(x^{S=i_k}_{i};\theta_Q))}{\sum_{i_{lk}}h(f(x^{S=i_l}_{j};\theta_P), f(x^{S=i_k}_{i};\theta_Q))}.
\end{aligned}
\label{eq:p^do}
\end{equation}


\noindent \textbf{Overall Loss Function.}
Taking into account all the above discussions, the \emph{Causal-DFQ} loss can be calculated with differentiability and the overall loss function can be written as follows:
\begin{equation}
    \mathcal{L}_{overall} = \mathcal{L}_{vanilla}  + \lambda\cdot\mathcal{L}_{\emph{Causal-DFQ}},
\label{eq:overall_loss}
\end{equation}
where $\mathcal{L}_{\emph{vanilla}}$ is the objective from the vanilla data-free quantization loss, and $\lambda$ is the parameter to balance the targeted task and the distillation task. In practice, we adopt and modify the codebase of GDFQ~\cite{xu2020generative} to achieve our causality-based data-free quantization baseline, thus more details about the $\mathcal{L}_{\emph{vanilla}}$ can be found in GDFQ.

\subsection{Discussions on \emph{Causal-DFQ}}
Besides the derivation originated from the perspective of causality, we would like to give a straight-forward explanation of \emph{Causal-DFQ}. Combining Eq.\ref{eq:optimization} and Eq.\ref{eq:p^do}, we can observe that our method minimize the distributional distance between $P^{do(S=i_{lk})}(f(x_j;\theta_P) \mid f(x_i;\theta_Q)) $ and $P^{do(S=i_{qt})}(f(x_j;\theta_P) \mid f(x_i;\theta_Q))$. Specifically, with the critic function to estimate the conditional distribution, $P^{do(S=i_{lk})}(f(x_j;\theta_P) \mid f(x_i;\theta_Q)$ acts as the similarity matrix between generated images with same content, \textit{i.e.}, a series of differences among samples with the same content and different styles. Finally, the similarity matrices of pre-trained and quantized models are aligned with KL-divergence as shown in Fig.\ref{pipeline:2}. 

\noindent\textbf{Difference with RELIC~\cite{mitrovic2020representation}.}
From the perspective of causality, the most related work is RELIC~\cite{mitrovic2020representation} which acts as a regularizer in self-supervised learning via the independence mechanisms~\cite{peters2017elements} to encourage networks to be invariant to different augmentations of the same instance. 
This self-supervised learning method also constructs a causal graph to model the data generation process. However, the focus of this work is on the content invariant property using data augmentations to stimulate inaccessible interventions~\cite{zhang2021adversarial}, which varies from our data-free work, \emph{Causal-DFQ}. 
Specifically, our work is different from RELIC (and most of the previous causality-guided computer vision models, such as~\cite{xie2021unaligned,mitrovic2020representation,cheng2021style}) for two significant reasons. Firstly, there is a unique nature in data-free scenarios where both the content and style variable are accessible, and thus we do not need to stimulate the interventions on the style domain. Secondly, the derived equilibrium is different where we focus on the distributions of outputs of pre-trained and quantized models. Detailed differences between our data-free approach and previous works are discussed in Appendix.

\section{Experiments}

\begin{table*}[!h]\footnotesize
\centering
\caption{\textbf{Comparisons on ImageNet}. We quantize both the weights and activations of the models to \textit{6-bits} and report the top-1 accuracy.}
\scalebox{1.1}{
\begin{tabular}{c|c|cc|cccccc}\toprule
    \multirow{2}{*}{Dataset} & \multirow{2}{*}{Model}   & \multicolumn{2}{c|}{Real Data} & \multicolumn{5}{c}{Data Free} \\ \cline{3-10}
    & & FP32 & FT  & ZeroQ~\cite{cai2020zeroq}  & GDFQ~\cite{xu2020generative}  & DSG~\cite{zhang2021diversifying} & SQuant~\cite{guo2022squant} & IntraQ~\cite{zhong2022intraq} & \emph{Causal-DFQ}\\ \hline
             & ResNet-18    & 71.47    & 70.76  & 69.84 & 70.13 & 70.46 & 70.74 & 70.60  &   \textbf{71.01{\tiny~$\pm$~0.06}}   \\
             & ResNet-50    & 77.74    & 77.70  & 72.93 & 76.59 & 76.07 & 77.05 & 76.90  &   \textbf{77.45{\tiny~$\pm$~0.13}} \\
    ImageNet & Inception-v3 & 78.80    & 78.80  & 74.94 & 77.20 & -     & 78.30 & 77.46 &   \textbf{ 78.40{\tiny~$\pm$~0.02}} \\
             & SqueezeNext  & 69.38    & 68.78  & 16.54 & 65.46 & -     & 67.34 & 67.45  &   \textbf{ 67.87{\tiny~$\pm$~0.11}} \\
             & ShuffleNet   & 65.07    & 64.55  & 35.21 & 60.12 & -     & 60.25 & 60.18 &   \textbf{ 60.83{\tiny~$\pm$~0.06}} \\\bottomrule
\end{tabular}}
 \vspace{-0.1in}
\label{tabel:class_6bit}
\end{table*}

\begin{table*}[!h]\footnotesize
\centering
\caption{\textbf{Comparisons on CIFAR-10/100 and ImageNet with 4W4A quantization setting}.}
\scalebox{1.0}{
\begin{tabular}{c|c|cc|ccccccc}\toprule
    \multirow{2}{*}{Dataset} & \multirow{2}{*}{Model}   & \multicolumn{2}{c|}{Real Data} & \multicolumn{5}{c}{Data Free} \\ \cline{3-11}
    & & FP32 & FT & DFQ~\cite{nagel2019data} & ZeroQ~\cite{cai2020zeroq}  & GDFQ~\cite{xu2020generative}  & DSG~\cite{zhang2021diversifying} & SQuant~\cite{guo2022squant} & IntraQ~\cite{zhong2022intraq} & \emph{Causal-DFQ}\\ \hline
    CIFAR-10 & ResNet-20    & 94.03    & 93.11 & 89.03 & 79.30  & 90.25 & 78.99& - & 91.49  &   \textbf{92.30{\tiny~$\pm$~0.08}}  \\
    CIFAR-100& ResNet-20    & 70.33    & 68.34 & 63.21 & 45.20  & 63.58 & 46.03 & - & 64.98  &   \textbf{65.67{\tiny~$\pm$~0.28}}   \\\hline
             & BN-VGG16     & 74.28    & 68.83 & 45.56 & 1.15  & 67.10 & 31.06 & 68.32 & 68.73  &   \textbf{71.09{\tiny~$\pm$~0.30}}   \\
             & ResNet-18    & 71.47    & 67.84 & 55.78 & 26.04 & 60.60& 34.53 & 66.14 & 66.47  &   \textbf{68.11{\tiny~$\pm$~0.17}}   \\
    ImageNet & ResNet-50     & 77.74    & 72.89 & 47.34 & - & 70.23 & - & 70.80 & 70.65  &   \textbf{72.49{\tiny~$\pm$~0.22}} \\
             & Inception-v3 & 78.80    & 73.80 & 49.62 & 26.84 & 70.39 & 34.89 & 73.26 & 73.12  &   \textbf{73.35{\tiny~$\pm$~0.42}} \\
             & SqueezeNext  & 69.38    & 65.78 & - & - & 39.18 & - & 43.45 & 42.78  &  \textbf{45.99{\tiny~$\pm$~0.13}} \\\bottomrule
\end{tabular}}
 \vspace{-0.2in}
\label{tabel:class_4bit}
\end{table*}

\subsection{Experimental Setup}
\noindent\textbf{Datasets.} 
We validate the \emph{Causal-DFQ} on four well-known data sets including CIFAR-10, CIFAR-100~\cite{krizhevsky2009learning}, ImageNet~\cite{deng2009imagenet} for recognition, and PASCAL VOC 2012~\cite{everingham2015pascal} for detection. 
More details about the datasets are in Supplemental Materials.

\noindent\textbf{Baselines.}
To evaluate the effectiveness and advantages of our proposed method, we compared it with both data-free fine-tuning methods and post-training quantization methods. The baselines are presented as follows. 
\textbf{FP32}: the full-precision pre-trained model. \textbf{FT}: we use real training data instead of fake data to fine-tune the quantized model by minimizing L2. \textbf{ZeroQ}~\cite{cai2020zeroq}: a data-free post-training quantization method. \textbf{DFQ}~\cite{nagel2019data}: a post-training quantization method uses a weight equalization scheme to remove outliers in both weights and activations. \textbf{ZAQ}~\cite{liu2021data}: It is a fine-tuning method by optimizing the quantized models in an adversarial learning way. \textbf{DSG}~\cite{zhang2021diversifying}: It is a fine-tuning method where the diversity of generated data is enhanced. \textbf{GDFQ}~\cite{xu2020generative}: It is also a fine-tuning method for recovering fake data via a conditional generator. \textbf{SQuant}~\cite{guo2022squant} and \textbf{IntraQ}~\cite{zhong2022intraq} are recently SoTA. Note that our code is modified from the code of GDFQ. 

\noindent\textbf{Implementation Details.} 
On CIFAR, we optimize the generator and quantized model using Adam~\cite{kingma2014adam} and SGD with Nesterov~\cite{nesterov1983method} respectively, where
the momentum term and weight decay in Nesterov are set to $0.9$ and $1\times{10}^{-4}$. Moreover, the learning rates of quantized models and generators are initialized to $1\times{10}^{-4}$and $1\times{10}^{-3}$ respectively. Both of them are decayed by $0.1$ for every 100 epochs. In addition, we train the generator and quantized model for 400 epochs with 200 iterations per epoch. On ImageNet, we set the initial learning rate of the quantized model as $1\times{10}^{-6}$. Other training settings are the same as those on CIFAR.
More details can be found in \emph{Supplemental Materials}.

\subsection{Comparison to SoTA}

\noindent\textbf{Image Classification.} We quantize both weights and activations to 6-bit, and report the comparison results in Table~\ref{tabel:class_6bit}. We also quantize them to 4-bit, and report the results in Table~\ref{tabel:class_4bit}. In all three classification datasets, our method \emph{Causal-DFQ} outperforms other existing state-of-the-art methods with various network architectures. In particular, when the number of categories increases in CIFAR-100, our method suffers a much smaller accuracy degradation than other methods. 
The main reason is that our method based on causality gains more prior knowledge from the full-precision model. These results demonstrate the superiority of our method. Especially for the large-scale dataset, ImageNet, existing data-free quantization methods suffer from severe performance degradation. 
However, our generated images contain style-irrelevant information and satisfy the similar distribution of real data. As a result, our method recovers the accuracy of quantized models significantly with the help of the content-style-decoupled generator and style-intervened discrepancy reduction on three commonly-used networks. 

\begin{figure}[t!]
  \begin{center}
    \includegraphics[width=0.48\textwidth]{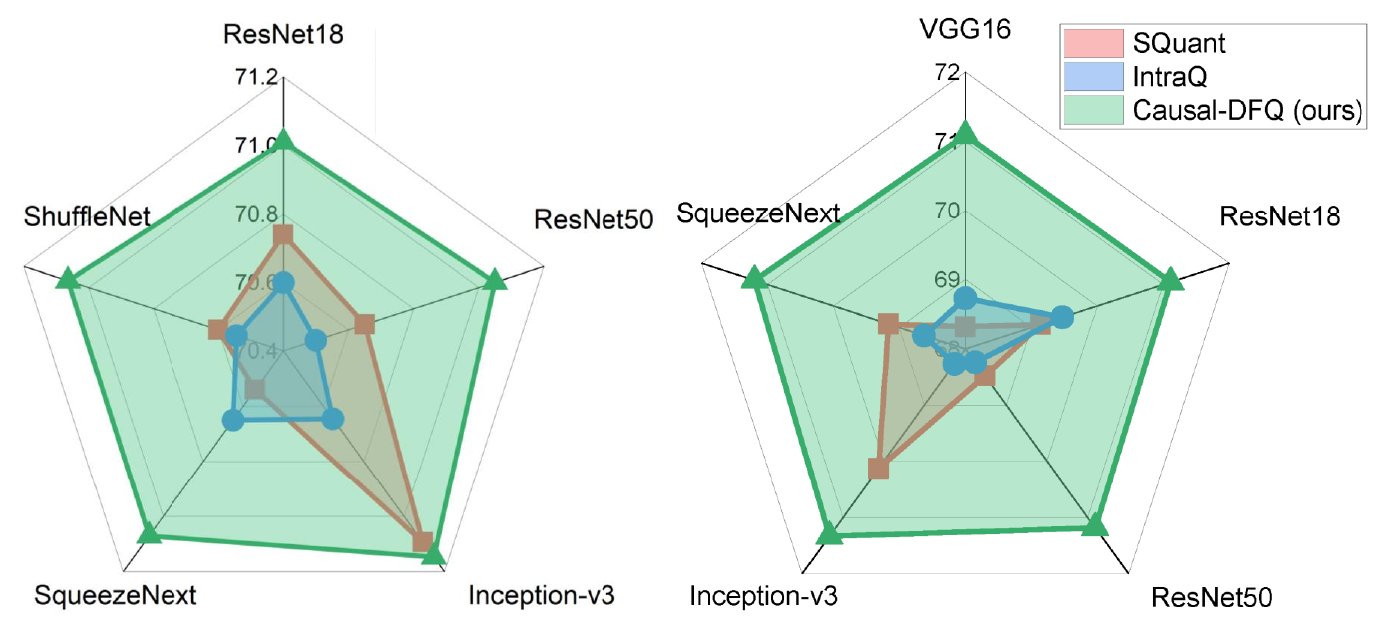}
  \end{center}
  \vspace{-0.1in}
  \caption{Overall performance on 6-bit (Left, corresponding to Tab.~\ref{tabel:class_6bit}) and 4-bit (Right, corresponding to Tab.~\ref{tabel:class_4bit}) settings.}
  \label{fig:rader_performance}
\end{figure}

Importantly, \emph{Causal-DFQ} comprehensively outperforms recent SoTA~\cite{guo2022squant} and~\cite{zhong2022intraq} as shown in Fig.~\ref{fig:rader_performance}. 
There is a breakthrough where the data-free quantized models fine-tuned by \emph{Causal-DFQ} outperform the (quantized) ones re-trained with real data \wrt accuracy on the ImageNet dataset. In addition, data-free quantization is more efficient in terms of training time, \eg, fine-tuning 4-bit ResNet via our data-free quantization method costs 8.4 GPU hours while re-training in a data-given manner costs 29.6 hours. These experimental results demonstrate that data-free quantization can empirically replace the method of re-training low-bit networks.

\begin{table}[th]
    \centering
    \caption{\textbf{Comparisons on VOC 2012 for object detection}. mAP is the metric, and higher is better.}
    \scalebox{0.95}{
    \begin{tabular}{p{2.0cm}|cccc}\toprule
   \scalebox{0.8}{\diagbox{Method}{Bits}} & W8A8 & W4A8 & W4A4 & W2A2 \\\hline
   FT & 70.35 & 68.24 & 64.28 & 57.12 \\\hline
   DFQ~\cite{nagel2019data} & 69.16 & 64.57 & 13.15 & 2.65 \\
   ZeroQ~\cite{cai2020zeroq} & 69.04 & 67.53 & 62.72 & 56.96 \\
   ZAQ~\cite{liu2021zero} & 70.02 & 68.12 & 64.44 & 56.96 \\
   Ours & \textbf{70.63} & \textbf{68.45} & \textbf{66.10} & \textbf{57.26} \\\bottomrule
\end{tabular}}
\label{tabel:detection}
\end{table}

\begin{figure*}[t!]
    \centering
    \includegraphics[width=0.9\textwidth]{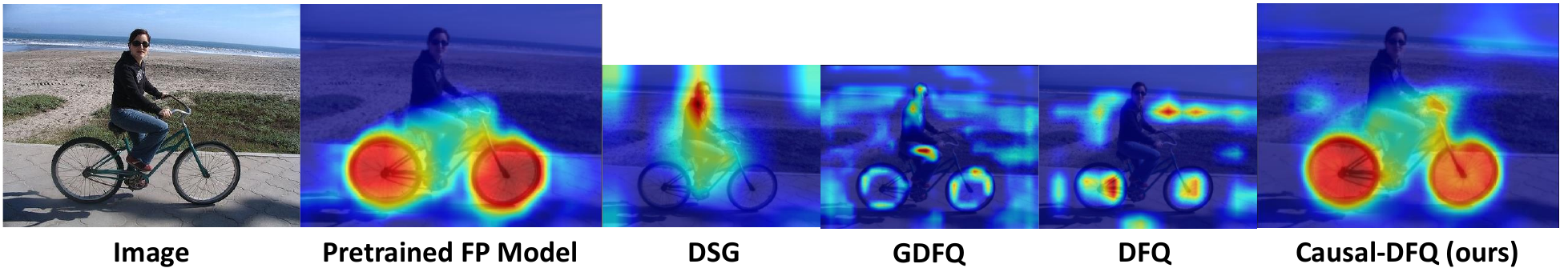}
    \vspace{-0.05in}
    \captionsetup{font=small}
    \caption{What makes the data-free quantized network for detection on VOC think the pixel label is `bicycle', visualized via Grad-Cam~\cite{selvaraju2017grad}. We can see that model quantized by \emph{Cuasal-DFQ} is able to focus on task-specific Content.}
    \label{fig:grad-cam}
\vspace{-0.2in}
\end{figure*}

\begin{figure}[t!]
  \begin{center}
    \includegraphics[width=0.42\textwidth]{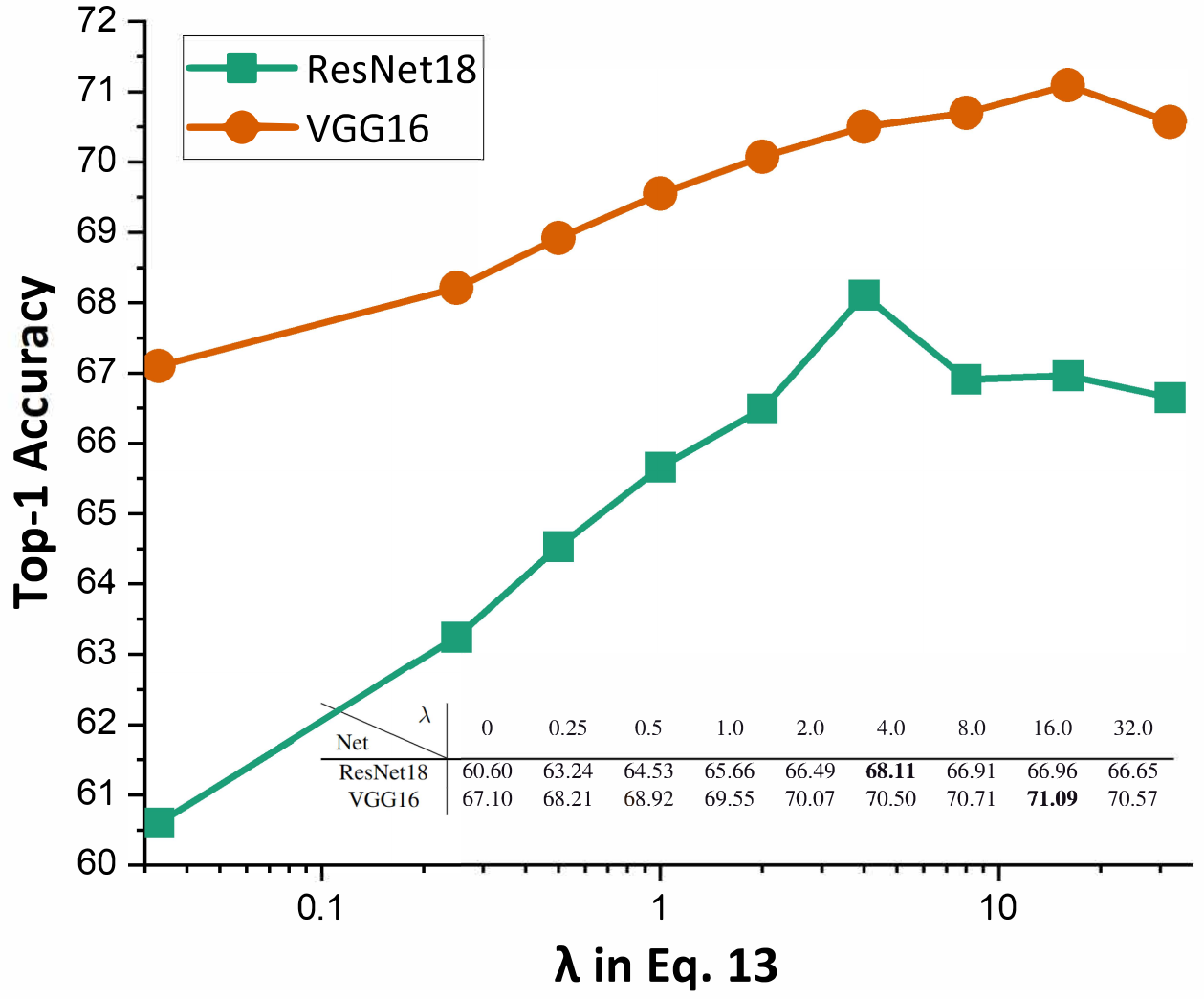}
  \end{center}
  \vspace{-0.1in}
  \caption{Ablation Study: Effect of $\lambda$. Note that $\lambda = 0$ equals to no \emph{Causal-DFQ} as our baseline (\ie, GDFQ~\cite{xu2020generative}).}
  \label{fig:ablation}
  \vspace{-0.2in}
\end{figure}

\noindent\textbf{Object Detection.} To demonstrate the application on object detection, we apply \emph{Causal-DFQ} to the model MobileNetV2 SSD~\cite{liu2016ssd} and evaluate it on VOC2012. Table~\ref{tabel:detection} demonstrates the advantages of our method compared to other quantization methods. In particular, \emph{Causal-DFQ} also outperforms FT that utilizes the original training dataset.

\subsection{Ablative Studies and analyses}

\noindent\textbf{Ablation Study.}\par
We conducted a series of ablative studies of our proposed method on ImageNet with the ResNet18 and VGG16 architectures. By adjusting the coefficient $\lambda$ in the loss function (Eq.\ref{eq:overall_loss}), where $\lambda = 0$ equals to no \emph{Causal-DFQ} as our baseline (\ie, GDFQ~\cite{xu2020generative}). The results are shown in Fig.\ref{fig:ablation}. With $\lambda$ increasing, the performance improvements show the effectiveness of our method. However, when the ratio of $\mathcal{L}_{\emph{Causal-DFQ}}$ in $\mathcal{L}_{overall}$ (Eq.~\ref{eq:overall_loss}) is greater than 10\% (on average), data-free quantization performance drops. 
A well-trained quantized network should have both the ability to align low-level feature maps (\ie, aligning as GDFQ~\cite{xu2020generative}) and learn from causality (\ie, \emph{Causal-DFQ}).




\noindent\textbf{Network Similarity between FP and Quantized Networks.}\par
Centered kernel alignment (CKA)~\cite{cortes2012algorithms,tung2019similarity} analyzing (hidden) layer representations of neural networks, enabling quantitative comparisons of representations within and across networks. It is a widely acknowledged tool for measuring the similarity between two networks~\cite{raghu2021vision}. Higher similar score between two layers' output representations mean those two layers share more similarity. The visualization of CKA analysis is presented in Fig.~\ref{fig:similarity}. More details about CKA for metricing network similarity are in \emph{Supplemental Materials}. 
\begin{figure}[h!]
  \begin{center}
    \includegraphics[width=0.49\textwidth]{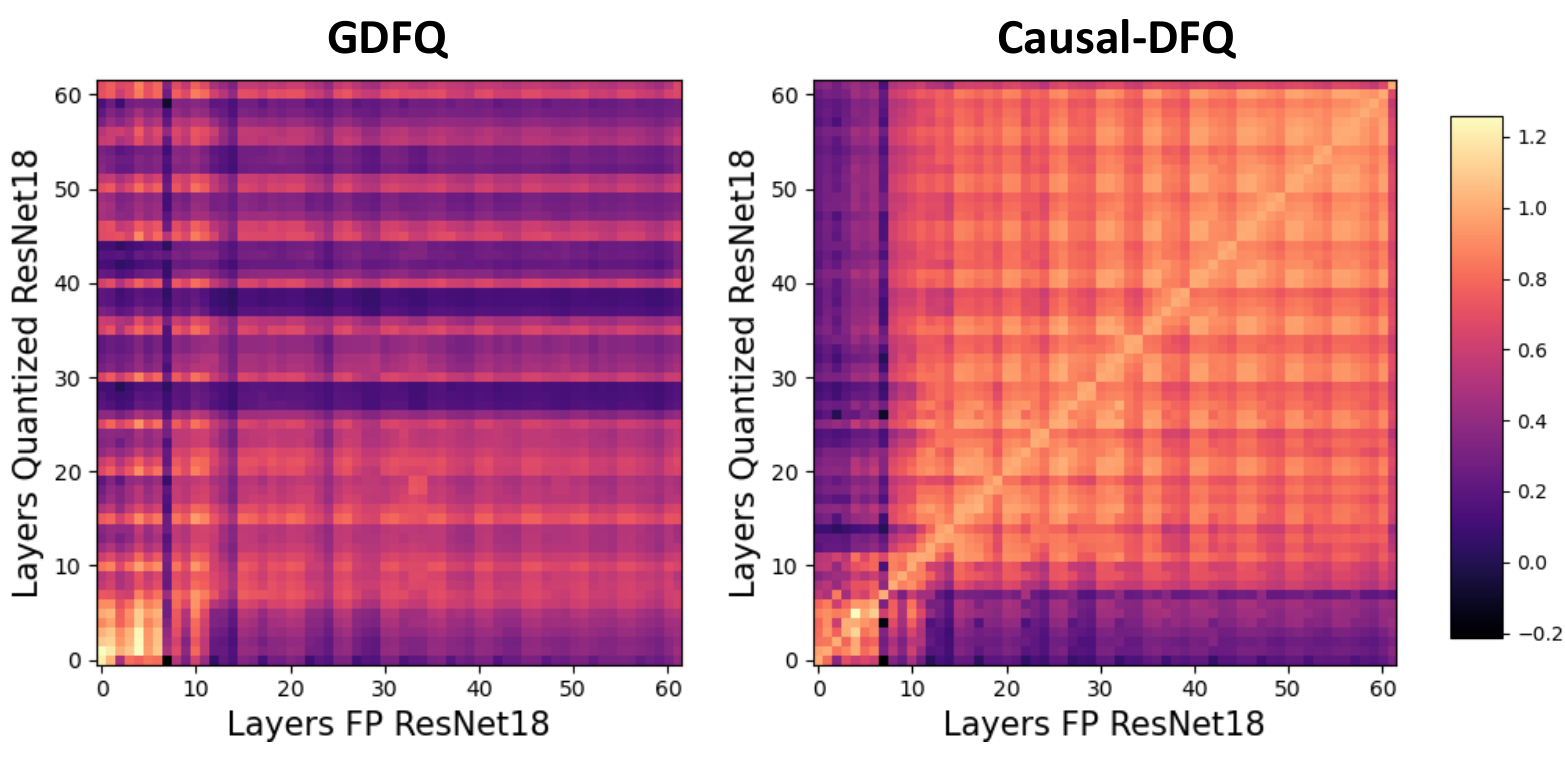}
  \end{center}
  \vspace{-0.1in}
  \caption{Cross model CKA~\cite{cortes2012algorithms,tung2019similarity} heatmaps between FP and quantized networks. The lighter the dot, the more similar of the two corresponding layers learned from different datasets. We can conclude that quantized network trained by \emph{Causal-DFQ} is more similar to the FP network.}
  \label{fig:similarity}
  \vspace{-0.2in}
\end{figure}

\noindent\textbf{Attention of Quantized Model Analysis via Grad-Cam~\cite{selvaraju2017grad} Visualization}.\par
We analyze the attentions of several quantized models \wrt targeted task. The results are presented in~\ref{fig:grad-cam}. We can see that the quantized model created by our method behaves more similarly to the pre-trained model. 
Thus, we conclude that \emph{Causal-DFQ} can quantized pre-trained FP model in a content-preserving manner.

\section{Conclusion}
\label{sec:con}
In this paper, we introduce causal reasoning into data-free quantization. We first formalize a causal graph to model the data-free quantization mechanism. Based on the causal graph, we propose the Causality-guided Data-free Network Quantization method to eliminate the reliance on data while training a quantized model. Specifically, we design a generator which can generate images conditioned on the content and style variables in the view of causality, and then we devise a discrepancy reduction loss to align the intervened distributions of the outputs of pre-trained and quantized models. 

\noindent\textbf{Acknowledgement.} This research was supported by the gift donation from Cisco. This article solely reflects the opinions of its authors and not the funding agent.
\clearpage

\section{Appendix}

\subsection{Experimental Setup}
\noindent\textbf{Datasets.} 
We validate the \emph{Causal-DFQ} on four well-known data sets including CIFAR-10, CIFAR-100~\cite{krizhevsky2009learning}, ImageNet~\cite{deng2009imagenet} for recognition, and PASCAL VOC 2012~\cite{everingham2015pascal} for detection. Specifically, CIFAR-10 consists of 60k images from 10 classes, with 6k per class. There are 50k images for training and 10k images for testing. CIFAR-100 has 100 classes, and each class contains 500 training images and 100 testing images. ImageNet is one of the most challenging and largest benchmark datasets for image classification, which has around 1.2 million real-world images for training and 50k images for validation. VOC 2012 contains 11,540 images, and each image contains a set of objects out of 20 different classes. 

\noindent\textbf{Implementation Details.} 
On CIFAR, we optimize the generator and quantized model using Adam~\cite{kingma2014adam} and SGD with Nesterov~\cite{nesterov1983method} respectively, where
the momentum term and weight decay in Nesterov are set to $0.9$ and $1\times{10}^{-4}$. Moreover, the learning rates of quantized models and generators are initialized to $1\times{10}^{-4}$and $1\times{10}^{-3}$ respectively. Both of them are decayed by $0.1$ for every 100 epochs. In addition, we train the generator and quantized model for 400 epochs with 200 iterations per epoch. On ImageNet, we set the initial learning rate of the quantized model as $1\times{10}^{-6}$. Other training settings are the same as those on CIFAR.
More implementation details can be found in the \textbf{codes}.

\subsection{Causal Reasoning}
\label{related_causal}
One core purpose of causal reasoning is to pursue the causal effect of interventions, contributing to achieving the desired objective. Recent work shows the benefits of introducing causality into machine learning from various aspects. After the deep connections of causal systems and the concept of exogeneity having been successfully implemented in social science, such as in Economics and Genetics~\cite{pearl2009causality}, Sch{\"o}lkopf \textit{et al.}~\cite{scholkopf2012causal} originally develop a technique, named independence mechanisms via introducing casual mechanisms to independently separate the exogenous and endogenous variables \textit{w.r.t.} specific tasks in the field of machine learning~\cite{scholkopf2021toward}. Specifically, given two variables $C$ and $E$, if $P(E\mid C)$ remains invariant to changes in the process that generates $C$, then $C$ can be defined as an exogenous variable. Huang \textit{et al.}~\cite{huang2020causal} also prove that $P(C)$ and $P(E\mid C)$ change independently of each other when they both change. Besides the theoretical studies, there are several works implementing this mechanism into other tasks for not only improving the model performance but also understanding the tasks from the perspective of casual inference. For example, Mitrovic \textit{et al.}~\cite{mitrovic2020representation} propose to enforce invariant prediction \textit{w.r.t.} style changes through an invariance regularizer, which yields improved generalization guarantees in self-supervised learning. Xie \textit{et al.}~\cite{xie2021unaligned} bridge the causally independent hypothesis with the image-to-image translation. Chen \textit{et al.}~\cite{chen2021style} reveal that intra-domain style invariance is also of pivotal importance to improve domain generalization approaches. Apart from the independence mechanism, causal reasoning is also introduced to several CV fields~\cite{scholkopf2019causality}, such as long-tail recognition~\cite{tang2020long}, semantic segmentation~\cite{zhang2020causal}, few-shot learning~\cite{yue2020interventional} and class-incremental learning~\cite{hu2021distilling}.

\textbf{The most related work is RELIC}~\cite{mitrovic2020representation} which acts as a regularizer in self-supervised learning via the independence mechanisms~\cite{peters2017elements} to encourage networks to be invariant to different augmentations of the same instance. This self-supervised learning method also constructs a causal graph to model the data generation process. However, the focus of this work is on the content invariant property using data augmentations to stimulate inaccessible interventions~\cite{zhang2021adversarial}. However, our work is different from RELIC for two significant reasons. 
In summary, thanks to the unique nature of data-free quantization, our data generation process is steerable, unlike previous works. 
Thus, we can design a content-style-decoupled generator where both content and style variables are accessible in the causal graph. Then we can easily implement do-calculus~\cite{pearl2012calculus} for causal reasoning. 
In the following section, we will discuss the differences in causal literature.

\subsection{Previous Causality-based Methods.}
Existing works~\cite{scholkopf2012causal,huang2020causal,mitrovic2020representation,xie2021unaligned} on different tasks choose the invariant content $C$ rather than the whole data $X$ as the optimal node to extract representations. Since the content $C$ is an informative representation for the data $X$ and inference based on $C$ is more stable against perturbations due to its causal dependency with the ground truth. Therefore, they expect that the information represented and extracted from data to the network $f(\cdot;\theta)$ should be invariant to style and correlated to content. We theoretically interpret this idea as follows: $\forall i_l, i_k \in \mathcal{I}$
\begin{equation}
    P^{do(S=i_l)}(\tilde{Y}\mid f(\tilde{X};\theta)) = P^{do(S=i_k)}(\tilde{Y}\mid f(\tilde{X};\theta)) ,
\label{eq:goal1}
\end{equation}
where $f(\cdot;\theta)$ is a model with parameters $\theta$.

However, the style domain $\mathcal{S}$ is practically inaccessible in the most learning scenarios such as supervised learning~\cite{huang2020causal,scholkopf2012causal} and self-supervised learning~\cite{mitrovic2020representation,xie2021unaligned,cheng2021style}, it is quite challenging to perform interventions on styles. \textbf{To stimulate the intervention}s on the style domain $\mathcal{S}$, previous works~\cite{xie2021unaligned,mitrovic2020representation,cheng2021style} \textbf{adopt the augmentation operations on data domain} $\mathcal{X}$.
Specifically, the data augmentations (\textit{e.g.}, Gaussian blurring, flips, rotation, color distortions, and random cropping) on data $X$ are utilized as interventions on the style variable $S$. Hence, the interventions $i_l$ and $i_k$ in the do-calculus of Eq.~\ref{eq:goal1} can be replaced by augmentations $a_l$ and $a_k$, and then we can derive the representation of invariant content by student networks. 
This relaxed equilibrium can be written as follows: $\forall l, k \in \left\{1,2,\cdots,M_A\right\}$,
\begin{equation}
\begin{aligned}
    P^{do(X=a_l)}(\tilde{Y}\mid f(\tilde{X};\theta)) &= P^{do(X=a_k)}(\tilde{Y}\mid f(\tilde{X};\theta))
\end{aligned}
\label{eq:realgoal_0}
\end{equation}
where $a_l$ and $a_k$ form a pair of augmentations in the domain of interventions $\mathcal{A}$, $M_A$ is the number of augmentations in $\mathcal{A}$, and $P^{do(X=a_l)}$ stands for the distribution under augmentation $a_l$ on $\mathcal{X}$~\cite{mitrovic2020representation}. 
Then, the targeted representation of invariant content of the network in Eq.\ref{eq:goal1} can be obtained by Eq.\ref{eq:realgoal_0}. Finally, the causal dependency between invariant $C$ and the output of network $f(\cdot;\theta)$ is discovered. 
\noindent\textbf{Difference with Previous Works.}
In our data-free quantization scenario, via generating mechanism $\mathcal{M}$ in Eq.4 (Sec.3.2), we can access style domain $\mathcal{S}$ and content domain $\mathcal{C}$. On the contrary, in most machine learning scenarios, content variable is computationally inaccessible as they use natural data, whose content variable is ill-defined. 

Specifically, in the content of data-free quantization, we successfully escape from the aforementioned awful predicament where the style domain $\mathcal{S}$ is practically inaccessible. Thus we can directly operate the style domain $\mathcal{S}$ and obtain corresponding distribution $P^{do(S=i_l)}(f(\tilde{X};\theta_P)\mid f(\tilde{X};\theta_Q))$ as formulated in \textbf{Targeted Causal Equilibrium} (Eq.7 in Sec.3.2)
\begin{equation}
\begin{aligned}
     &P^{do(S=i_l)}(f(\tilde{X};\theta_P)\mid f(\tilde{X};\theta_Q))\\
    =&P^{do(S=i_k)}(f(\tilde{X};\theta_P)\mid f(\tilde{X};\theta_Q)),
\end{aligned}
\label{eq:realgoal}
\end{equation}
where $M$ is the number of interventions in style domain $\mathcal{S}$, $f(\cdot;\theta_P)$ and $f(\cdot;\theta_Q)$ are the pre-trained and quantized model with parameters $\theta_P$ and $\theta_Q$, respectively. 
Straight-forwardly, we desire the distribution, $P(f(\tilde{X};\theta_P)\mid f(\tilde{X};\theta_Q))$ to be invariant over style variable change.

{\small
\bibliographystyle{ieee_fullname}
\bibliography{egbib}
}

\end{document}